\documentclass[10pt,twocolumn,letterpaper]{article}

\usepackage{cvpr}
\usepackage{times}
\usepackage{epsfig}
\usepackage{graphicx}
\usepackage{amsmath}
\usepackage{amssymb}

\usepackage[capbesideposition=outside,capbesidesep=quad]{floatrow}

\usepackage{xr} 
\makeatletter 
\newcommand*{\addFileDependency}[1]{%
  \typeout{(#1)} 
  \@addtofilelist{#1} 
  \IfFileExists{#1}{}{\typeout{No file #1.}} 
} 
\makeatother 
 
\newcommand*{\myexternaldocument}[1]{%
    \externaldocument{#1}%
    \addFileDependency{#1.tex}%
    \addFileDependency{#1.aux}%
} 
\myexternaldocument{supp}

\usepackage{titling}

\usepackage{enumitem}
\usepackage{amssymb}
\usepackage{amsmath}
\usepackage{mathtools}
\usepackage{graphicx}
\usepackage{dsfont}  %
\usepackage{booktabs} %
\usepackage{multirow}
\usepackage{algorithm}
\usepackage{algorithmic}

\usepackage{color}
\definecolor{atomictangerine}{rgb}{0.8, 0.2, 0.1}
\definecolor{turq}{rgb}{0.0, 0.5, 0.5}
\definecolor{darkturq}{rgb}{0.0, 0.4, 0.4}
\definecolor{bright}{rgb}{0.8, 0.1, 0}
\definecolor{darkgray}{gray}{0.3}
\definecolor{mahogany}{rgb}{0.6, 0.05, 0.05}
\definecolor{myblue}{rgb}{0.3,0.05,0.9}

\definecolor{olive}{rgb}{0.537, 0.627, 0.318}
\definecolor{green}{rgb}{0.22, 0.463, 0.114}
\definecolor{grey}{rgb}{0.4, 0.4, 0.4}
\definecolor{blue}{rgb}{0.435, 0.659, 0.863}
\definecolor{pink}{rgb}{0.761, 0.482, 0.627}
\definecolor{darkpink}{rgb}{0.561, 0.282, 0.427}

\newcommand\ignore[1]{}

\newcommand{\tildeapprox}{{\raise.17ex\hbox{$\scriptstyle\sim$}}}
\newcommand{\secref}[1]{Section \ref{#1}}
\newcommand{\figref}[1]{Figure \ref{#1}}

\renewcommand{\eqref}[1]{Eq.~(\ref{#1})}

\renewcommand\vec[1]{\mathbf{#1}}

\newcommand{\COSMO}{COSMO{} }
\newcommand{\COSMOLAGO}{COSMO+LAGO{} }
\newcommand{\COSMOGAN}{COSMO+fCLSWGAN{} }
\newcommand{\CSLAGO}{CS+LAGO{} }
\newcommand{\dazl}{\COSMO}
\newcommand{\DAZL}{\COSMO}

\newcommand{\gzsl}{generalized zero-shot learning{} }

\renewcommand{\a}{\vec{a}}
\newcommand{\x}{\vec{x}}
\newcommand{\y}{y}

\renewcommand{\S}{\mathcal{S}}
\newcommand{\U}{\mathcal{U}}
\renewcommand{\H}{\mathcal{H}}
\newcommand*{\medcup}{\mathop{\mathbin{\scalebox{1.5}{\ensuremath{\cup}}}}}

\newcommand{\PriorOOD}{\pi^{\U}}
\newcommand{\PriorinD}{\pi^{\S}}

\newcommand{\topkT}{CB-Gating-3{} }
\newcommand{\topkO}{CB-Gating-1{} }
\newcommand{\topkNoZS}{CB-Gating-3 (w/o $p^{ZS}$){} }
\newcommand{\pmaxT}{Max-Softmax-3{} }
\newcommand{\pmaxO}{Max-Softmax-1{} }
\newcommand{\Acctr}{$Acc_{tr}$ }
\newcommand{\Accts}{$Acc_{ts}$ }
\newcommand{\AccH}{$Acc_H$ }

\newcommand{\ADmodel}{Gating}
\newcommand{\gating}{gating }
\newcommand{\Gating}{Gating }
\newcommand{\laplaceAd}{Adaptive-Smoothing }

\newcommand{\laplaceConst}{Const-Smoothing }
\newcommand{\pg}{p^{Gate}}

\usepackage[breaklinks=true,bookmarks=false]{hyperref}

\cvprfinalcopy %

\def\cvprPaperID{5671} %

\ifcvprfinal\fi
\begin{document}

\title{Adaptive Confidence Smoothing for Generalized Zero-Shot Learning}

\author{ {Yuval Atzmon} \\
 Bar-Ilan University, NVIDIA Research \\
 {\tt\small yuval.atzmon@biu.ac.il}
 \and
{Gal Chechik}  \\
 Bar-Ilan University, NVIDIA Research \\
 {\tt\small gal.chechik@biu.ac.il}
}
\predate{}
\postdate{}
\date{}

\maketitle

\begin{abstract}
Generalized zero-shot learning (GZSL)  is the problem of learning a classifier where some classes have samples and others are learned from side information, like semantic attributes or text description, in a zero-shot learning fashion (ZSL). Training a single model that operates in these two regimes simultaneously is challenging. 
Here we describe a probabilistic approach that breaks the model into three modular components, and then combines them in a consistent way. Specifically, our model consists of three classifiers: A ``gating" model that makes soft decisions if a sample is from a ``seen" class, and two experts: a ZSL expert, and an expert model for seen classes. 
We address two main difficulties in this approach:
How to provide an accurate estimate of the \emph{gating} probability without any training samples for unseen classes; and how to use expert predictions when it observes samples outside of its domain.

The key insight to our approach is to pass information between the three models  to improve each one's accuracy, while maintaining the modular structure. We test our approach, \emph{adaptive confidence smoothing} (\COSMO\!), on four standard GZSL benchmark datasets and find that it largely outperforms state-of-the-art GZSL models. \COSMO is also the first model that closes the gap and surpasses the performance of generative models for GZSL, even-though it is a light-weight model that is much easier to train and tune.
Notably, COSMO offers a new view for developing zero-shot models. Thanks to COSMO's modular structure,  instead of trying to perform well both on seen and on unseen classes, models can focus on accurate classification of unseen classes, and later consider seen class models.

\end{abstract}

\section{Introduction}
Generalized zero-shot learning (GZSL) \cite{chao} is the problem of learning to classify samples from two different domains of classes: {\em seen classes}, trained in a standard supervised way from labeled samples, and {\em unseen classes}, learned from external knowledge, such as attributes or natural language, in a zero-shot-learning fashion. GZSL poses a unique combination of hard challenges: First, the model has to learn effectively for classes without samples (zero-shot). It also needs to learn well for classes with many samples. Finally, the two very different regimes should be combined in a consistent way in a single model. GZSL can be viewed as an extreme case of classification with unbalanced classes, hence solving the last challenge can lead to better ways of addressing class imbalance, which is a key problem in learning with real-world data.

\begin{figure}[tbp]
    \centering
      \hspace*{-.15cm}
      \includegraphics[height=4cm, trim={3.5cm 5.2cm 3.5cm 5.15cm},clip]{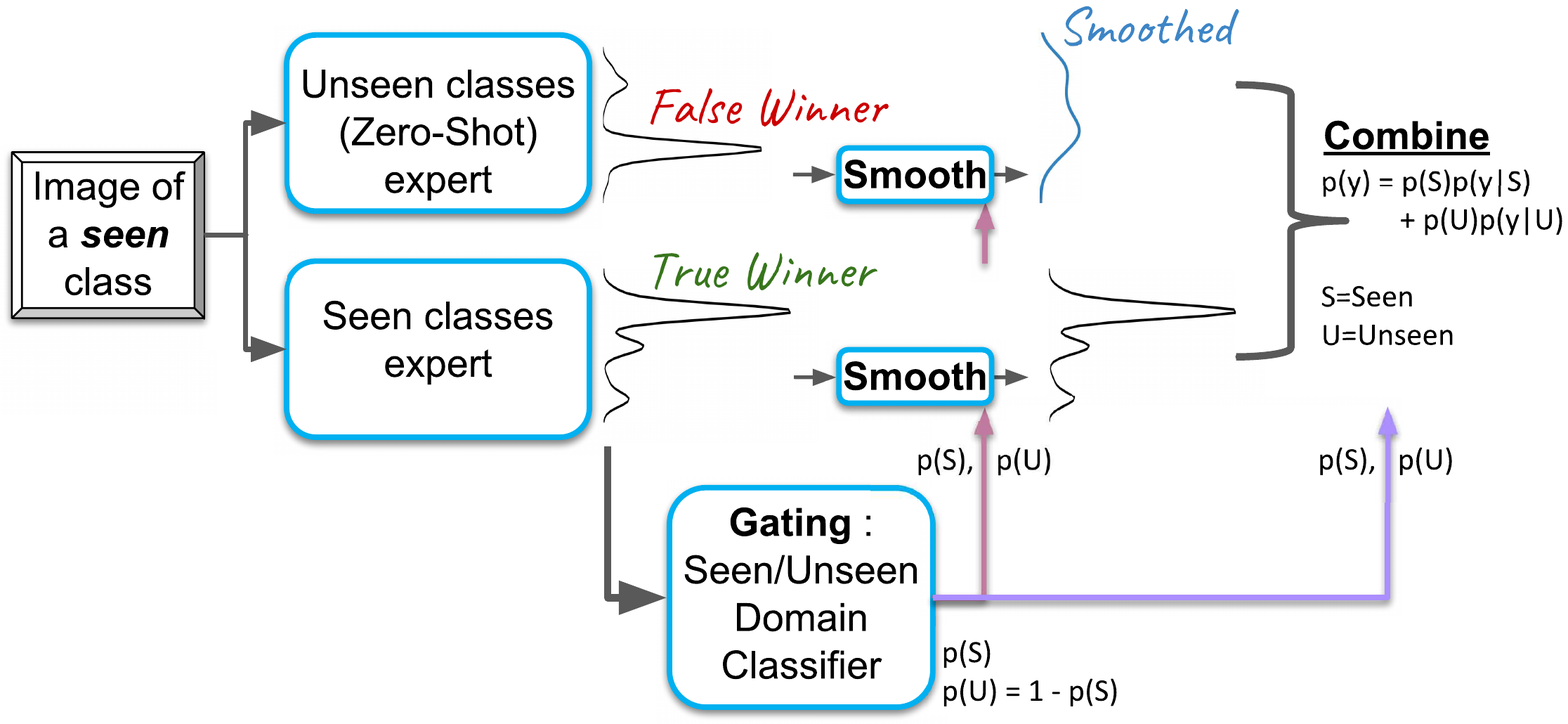} %
      \caption{A qualitative illustration of \COSMO: An input image is processed by two experts: A \textit{seen}-classes expert, and an \textit{unseen}-classes expert, which is a zero-shot model. (1) When an image is from a \textit{seen} class, the zero-shot expert may still produce an overly-confident false-positive prediction. \COSMO smooths the predictions of the unseen expert if it believes that the image is from a seen class. The amount of smoothing is determined by a novel gating classifier. (2) Final GZSL predictions are based on a soft combination of the predictions of two experts, with weights provided by the gating module.}
      \label{fig1}
\end{figure}

The three learning problems described above operate in different learning setups, hence combining them into a single model is challenging. Here, we propose an  architecture with three modules, each focusing on one problem. At inference time, these modules share their prediction confidence in a principled probabilistic way in order to reach an accurate joint decision. 

One natural instance of this modular architecture is \textit{hard gating}: Given a test sample, the gate assigns it either to a  \textit{seen expert}  - trained as a standard supervised classifier  - or to an \textit{unseen expert}  - trained in a zero-shot-learning fashion \cite{socher2013zero}. Only the selected expert is used for prediction, ignoring the other expert. Here we study a more general case, where both the seen expert and the unseen expert process each test sample, and their predictions are combined in a \textit{soft} way. Specifically, the predictions are combined by the soft gater using the law of total probability: $p(class) = p(class|seen)p(seen) + p(class|unseen)p(unseen)$.

Unfortunately, softly combining expert decisions raises several difficulties. First, when training a gating module it is hard to provide an accurate estimate of the probability that a sample is from the ``unseen" classes, because by definition no samples have been observed from those classes. 
Second, experts tend to behave in uncontrolled ways when presented with out-of-distribution samples, often producing confident-but-wrong predictions. As a result, when using a soft combination of the two expert models, the ``irrelevant" expert may overwhelm the decision of the correct expert. 

We address these issues in two ways. First, we show how to train a binary \gating mechanism to classify the Seen/Unseen domain based on the distribution of softmax class predictions. The idea is to simulate the softmax response to samples of unseen classes using a held-out subset of training classes, and represent expert predictions in a class-independent way. Second, we introduce a Laplace-like prior \cite{laplace_smoothing} over softmax outputs in a way that uses information from the gating classifier. This additional information allows the experts to estimate class confidence more accurately. 

This combined approach, named \textit{adaptive COnfidence SMOothing} (\COSMO), has significant advantages. It can incorporate any state-of-the-art zero-shot learner as a module, as long as it outputs class probabilities; It is very easy to implement and apply (code provided) since it has very few hyper-parameters to tune; Finally, it outperforms competing approaches on all four GZSL benchmarks (AWA, SUN, CUB, FLOWER).
Our main novel contributions are:
\vspace{-5pt}
\begin{itemize}[leftmargin=*]
    \setlength\itemsep{-0.4em}
    \item A new soft approach to combine decisions of seen and unseen classes for GZSL.
    \item A new ``out-of-distribution" (OOD) classifier to separate seen from unseen classes, and a negative result, showing that modern OOD classifiers have limited effectiveness on ZSL benchmarks.
    \item New state-of-the-art results for GZSL for all four main benchmarks, AWA, SUN, CUB and FLOWER. \COSMO is the first model that is comparable to or better than generative models of GZSL, while being easy to train.
    \item A characterization of GZSL approaches on the seen-unseen accuracy plane.
\end{itemize}

\begin{figure*}[htbp]

    \centering
  \includegraphics[height=4.9cm, trim={2.cm 8.5cm 2.1cm 6.1cm},clip]{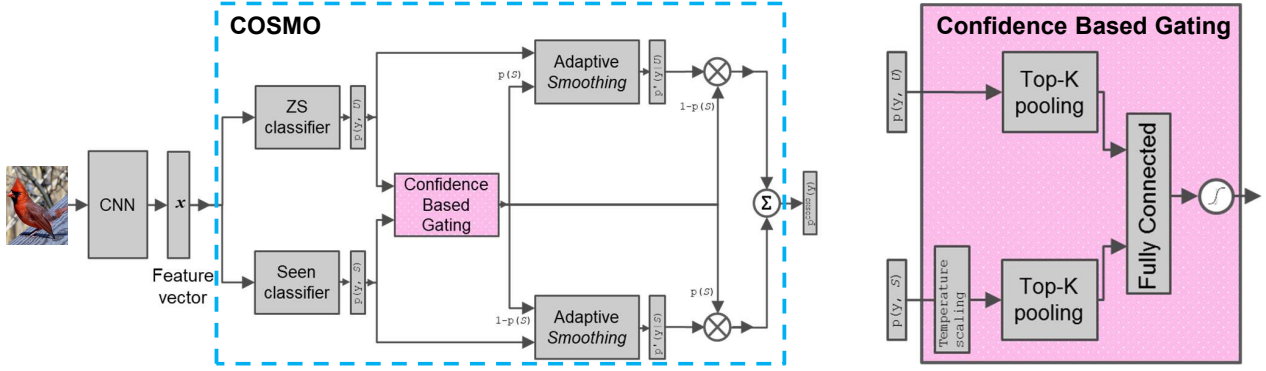} %
  \caption{
  \textbf{Left}, \COSMO Architecture: We decompose the GZSL task into three sub-tasks that can be addressed separately. (1) A model trained to classify seen  $\S$ classes. (2) A model classifying \textit{unseen}  $\U$ classes, namely a ZSL model, conditioned on $\U$. (3) A \textit{gating} binary classifier trained to discriminate between seen and unseen classes and to weigh the two models in a soft way;
  Before weighing (1) \& (2) softmax distributions, we add a prior for each if the gating network provides low confidence (Figure \ref{fig1} and Sec \ref{sec_domain_laplace_reg}).
    \textbf{Right}, The gating network (Zoom-in): It takes softmax scores as inputs. We train it to be aware of the response of softmax scores to \textit{unseen} images, with samples from held-out classes. Because test classes are different from train classes, we pool the top-$K$ scores, achieving invariance to class identity  (\secref{sec_domain_aware_gating}). The fully-connected layer only learns 10-50 weights (K is small) since this is a binary classifier.
     } 
     \label{fig2}
\end{figure*}

\section{Related work}
\label{sec_related_work}
In a broad perspective, \textbf{zero-shot learning} is a task of  \textit{compositional reasoning} \cite{Lake2014, Lake2017, atzmonComp, AndreasNAACL}, where new concepts are constructed by recombining primitive elements \cite{Lake2017}. This ability resembles human learning, where humans can easily recombine simple skills to solve new tasks \cite{Lake2014}.
ZSL has attracted significant interest in recent years \cite{xianCVPR, recentAdvances, DAP,  Rohrbach11, jayaraman2014zero, AlHalah15, DEM,  xu2017matrix, Li2018, Song_2018_CVPR}. As our main ZSL module, we use  \textit{LAGO} \cite{LAGO}, a state-of-the-art approach which learns to combine attribute that describes classes using an AND-OR group structure to estimate $p(class|image)$. 

\textbf{Generalized ZSL} extends ZSL to the more realistic scenario where the test data contains both seen and unseen classes. There are two kinds GZSL methods. First, some approaches synthesize feature vectors of unseen classes using generative models like VAE of GAN, and then use them in training \cite{xian_2018, CCGAN, CVAE2, CVAE1, ZhuGAN}. Second, approaches that use the semantic class descriptions directly during training and inference \cite{RelationNet, DEM, ICINESS, TRIPLE, DCN, socher2013zero, chao}.
To date, the first kind, methods that augment data, perform better. 

Among previous GZSL approaches, several are closely related to \COSMO\!. \cite{socher2013zero} uses a hard gating mechanism to assign a sample to one of two domain experts. \cite{ZhangGZSL} uses a soft gating mechanism and trains it using synthesized samples for unseen classes generated by a generative model \cite{xian_2018}. \cite{chao} calibrates between seen and unseen class scores by subtracting a constant value from seen class scores. \cite{DCN} uses temperature scaling \cite{hinton_temp} and an entropy regularizer to make seen class scores less confident and unseen scores more confident. 

\textbf{Detecting out-of-distribution samples:} 
Our approach to soft gating builds on developing an out-of-distribution detector, where \textit{unseen} images are treated as ``out-of-distribution'' samples. There is a large body of work on 1-class and anomaly detection which we do not survey here. 
In this context, the most relevant recent work includes \cite{Hendrycks, ODIN, Vyas, ShalevOOD}. \cite{Hendrycks} detects an OOD sample if the largest softmax score is below a threshold 
\cite{ODIN} scales the  softmax ``temperature''  \cite{hinton_temp} and perturbs the input with a small gradient step.  
\cite{ShalevOOD} represents each class by multiple word-embeddings and compares output norms to a threshold. \cite{Vyas} trains an ensemble of models on a set of ``leave-out" classes, with a margin loss that encourages high-entropy scores for left-out samples.

When testing  \cite{ODIN, Vyas} on the ZSL benchmarks studied in this paper (CUB, SUN, AWA), we found the perturbation approach of \cite{ODIN} hurts OOD detection, and that the loss of \cite{Vyas} overfits on the leave-out classes. We discuss possible explanation of these effects in Supplementary (\ref{sec_perturbation_results}).

\textbf{Mixture of experts (MoE):}
In MoE \cite{MoE_hinton, TwentyYearsMoE, sparse_deep_moe}, given a sample, a gating network first assigns weights to multiple experts. The sample is then classified by those experts, and their predictions are combined by the gating weights. All parts of the model are usually trained jointly, often using an EM approach. Our approach fundamentally differs from MoE in that, at training time, it is known for every sample whether or not it was already seen. As a result, experts can be trained separately without any need to infer latent variables, ensuring that each module is an expert of its own domain (seen or unseen).

\textbf{Cognitive psychology:} Our approach was inspired by dual-route models in cognitive psychology \cite{dualroute, fastslow}, where processing of information is performed by two cognitive systems: one is fast and intuitive for processing well known information, analogous to our ``seen'' expert. The second is slower and based on reasoning and analogous to our ``unseen'' expert.

\section{Generalized zero-Shot learning}
\label{sec:gzsl}
We start with a formal definition of zero-shot learning (ZSL) and then extend it to generalized ZSL.

In zero-shot learning, a training set $\mathcal{D}$ has $N$ labeled samples: $\mathcal{D} = \{ (\x_i, \y_i), i=1 \dots N \}$ , where each $\x_i$ is a feature vector and $\y_i\in\S$ is a label from a \textit{seen} class $\S = \{1,2, \dots |\S|\}$. 

At test time, a new set of samples $\mathcal{D}'=\{\x_i, i=N+1 \dots N+M\}$ is given from a set of \textit{unseen} classes $\U = \{|\S|+1, \dots |\S| + |\U|\}$. Our goal is to predict the correct class of each sample.
As a supervision signal, each class $y \in \S \medcup \U$ is accompanied with a  \textit{class-description} vector $\a_y$ in the form of semantic attributes \cite{DAP} or natural language embedding \cite{Reed, ZhuGAN, socher2013zero}. The crux of ZSL is to learn a compatibility score for samples and class-descriptions $F(\a_y, \x)$, and predict the class $y$ that maximizes that score.
In probabilistic approaches to ZSL \cite{DAP,lampert2014attribute, WangBN, socher2013zero,LAGO,DCN} the compatibility function assigns a  probability for each class $p(Y=y|\x)= F(\a_y, \x)$, with $Y$ viewed as a random variable for the label $y$ of a sample $\x$.

\textbf{Generalized ZSL:} While in ZSL test samples are drawn from the unseen classes $Y\in\U$, in GZSL samples are drawn from either the \textit{seen or unseen} domains: $Y\in\S \medcup \U$. 

\textbf{Notation:} Below, we denote an unseen class by $Y\in\U$ and a seen one by  $Y\in\S$. Given a sample $\x$ and a label $y$ we denote the conditional distribution that a class is seen by $p(\S)=p(Y\in\S|\x)$, or unseen $p(\U) = p(Y\in\U|\x) = 1 - p(Y\in\S|\x)$,  
and the conditional probability of a label by $p(y)=p(Y=y|\x)$, $p(y| \S)=p(Y=y| Y\in\S, \x)$ and $p(y| \U)=p(Y=y| Y\in\U, \x)$. For ease of reading, our notation does not explicitly state the conditioning on $\x$.

\section{Our approach}
\label{sec_our_approach}
We now describe COSMO, a probabilistic approach that breaks the model into three modules. The key idea is that these modules exchange information to improve each other's accuracy. Formally, by the law of total probability
\begin{eqnarray}
    p(y) &=& p(y| \S) p(\S) + p(y| \U) p(\U)\,.
    \label{eq_total_prob}
\end{eqnarray}
This formulation decomposes GZSL into three sub-tasks that can be addressed separately. (1) $p(y|\S)$ can be estimated by any model trained to classify seen $\S$ classes, whose prediction we denote by $p^{S}(y|\S)$. (2) Similarly, $p(y|\U)$ can be computed by a model classifying \textit{unseen}  $\U$ classes, namely a ZSL model, whose prediction we denote by $p^{ZS}(y|\U)$.
(3) Finally, the two terms are weighted by $p(\S)$ and $p(\U)=1-p(\S)$, which can be computed by a \textit{gating} classifier, whose prediction we denote by $\pg$, that is trained to distinguish seen from unseen classes. Together, we obtain a GZSL mixture model:
\begin{eqnarray}
    p(y) = p^{S}(y| \S) \pg(\S) + p^{ZS}(y| \U) \pg(\U) 
    \label{eq_total_prob_models}
\end{eqnarray}
A hard variant of \eqref{eq_total_prob_models} was introduced in \cite{socher2013zero}, where the gating mechanism makes a hard decision to assign a test sample to one of two expert classifiers, $p^{ZS}$ or $p^{S}$. 
Unfortunately, although conceptually simple, using a \textbf{soft} mixture model raises several problems. %

First, combining models in a soft way means that each model contributes its beliefs, even for samples from the other ``domain".
This tends to damage the accuracy because multiclass models tend to assign most of the softmax distribution mass to very few classes, even when their input is random noise \cite{Hendrycks}. For instance, when the unseen classifier is given an input image from a seen class, its output distribution tends to concentrate on a few spurious classes. This peaked distribution ``confuses" the combined GZSL mixture model, leading to a false-positive prediction of the spurious classes.
A second challenge for creating a soft gating model is to assign accurate weights to the two experts. This is particularly complex when discriminating seen from unseen classes, because it requires access to training samples of the unseen domain.

COSMO addresses these two problems using a novel confidence-based \gating network and by applying a novel prior during inference. Its inference process is summarized in Algorithm \ref{alg}, and a walk-through example is provided in Supplementary \ref{sec_walkthrough_example}. Next, we describe COSMO in detail. 

\subsection{Confidence-based \gating model}
\label{sec_domain_aware_gating}
The \gating module aims to decide if an input image comes from a seen class or an unseen class. 
Since no training samples are available for  unseen classes, we can view this problem in the context of \textit{Out-Of-Distribution} detection  by treating seen-class ($\S$) images as  ``in-distribution'' and unseen-class ($\U$) images as ``out-of-distribution''.

Several authors proposed training an OOD detector on in-distribution data, and detect an image as out-of-distribution if the largest softmax score is below a threshold \cite{Hendrycks,ODIN,Vyas}. 
Here we improve this approach by training a network on top of the softmax output of the two experts, with the goal of discriminating $\U$ images from  $\S$ images. Intuitively, this can improve the accuracy of the gating module because the output response of the two experts differs for $\S$ images and $\U$ images. We name this network as \textit{confidence-based gate} (CBG). It is illustrated in Figure \ref{fig2}.

One important technical complication is that training the CBG cannot observe any $\U$ images, because they must be used as unseen. We therefore create a hold-out set from $\S$  classes that are not used for training and use them to estimate the output response of the experts over $\U$ images. Below, we refer to this set of classes as held-out $\H$ classes, and their images as $\H$ images. Note that due to similar reasons we cannot train the gater jointly with the $\S$ and $\U$ experts. See Supplementary \ref{sec_joint_training} for details. 

This raises a further complexity: Training the unseen expert on $\H$ classes means that at test time, when presented with \textit{test} classes, The unseen expert should have an output layer that is different from its output layer during training. Specifically, it corresponds to new (test) classes, possibly with a different dimension. To become invariant to identity and the  order of $\H$ classes in the output of the expert, the CBG takes the top-$K$ scores of the soft max and sorts them. This process, which we call \textit{top-k pooling}, guarantees that the CBG is invariant to the specific classes presented. %
Top-K pooling   generalizes max-pooling, and becomes equivalent to max pooling for K=1. 

\subsection{Adaptive confidence smoothing}
\label{sec_domain_laplace_reg}

As we described above, probabilistic classifiers tend to assign most of the softmax mass to very few classes, even when a sample does not belong to any of the classes in the vocabulary. Intuitively, when given an image of out-of-vocabulary class as input, we would expect all classes to obtain a uniformly low probability, since they are all ``equally wrong".
To include this prior belief in our model we borrow ideas from Bayesian parameter estimation. Consider the set of class-confidence values  as the quantity that we wish to estimate, based on the confidence provided by the model (softmax output scores).  
In Bayesian estimation, one combines the data (here, the predicted confidence) with a prior distribution (here, our prior belief). 

Specifically, for empirical categorical (multinomial) data, \textit{Laplace smoothing} \cite{laplace_smoothing} is a common technique to achieve a robust estimate with limited samples. It amounts to adding ``pseudo counts" uniformly across all classes, and functions as a prior distribution over classes. We can apply a similar technique here, and combine the predictions with an additive prior distribution $\PriorOOD =p_0(y|\U)$. This yields 
\begin{align}
    \MoveEqLeft[3] p^{\lambda}(y|\U) = (1\!\!-\!\!\lambda) \, p(y| \U) + \lambda\,\PriorOOD    \quad, 
    \label{eq_da_laplace_lambda}
\end{align}
where $\lambda$ weighs the prior, and $\PriorOOD$ is not conditioned on $\x$. Similarly, for the seen distribution, we set $p^{\lambda}(y|\S) =  (1-\lambda) p(y|\S) + \lambda\PriorinD$. When no other information is available we set the prior to the maximum entropy distribution, which is the uniform distribution $\PriorOOD=1/{(\#\text{unseen classes})}$ and $\PriorinD=1/{(\#\text{seen classes})}$.

\paragraph{An Adaptive Prior: } %
How should the prior weight $\lambda$ be set? In Laplace smoothing, adding a constant pseudocount has the property that its relative weight decreases as more samples are available. Intuitively, this means that \textbf{when the data provides strong evidence, the prior is weighted more weakly}. We adopt this intuition for making the trade-off parameter $\lambda$ adaptive.
Intuitively, if we believe that a sample \textit{does not} belongs to a seen class, we smooth the seen classifier outputs (Figure \ref{fig1}).
More specifically, 
we apply an adaptive prior by replacing the constant $\lambda$ with our belief about each domain (for $p'(y|\U)$ set $\lambda = p(\U)$): 
\begin{eqnarray}
    p'(y|\U) &=& p(\U)p(y|\U) + (1\!\!-\!\!p(\U))\PriorOOD  \notag \\
    {} & = &p(y, \U) + (1\!\!-\!\!p(\U))\PriorOOD \quad. 
    \label{eq_da_laplace}
\end{eqnarray}
Similarly  $p'(y|\S) = p(\S)p(y|\S) + (1-p(\S))\PriorinD$.
In practice, we use the ZS model estimation for $p(y,\U)$, and the gating model estimation for $p(\U)$, yielding $p'(y|\U) = p^{ZS}(y, \U) + (1-\pg(\U))\PriorOOD$ and similarly for $p'(y|\S)$.
\ignore{
\begin{align}
    \MoveEqLeft[3] p'(y|\U) = p^{ZS}(y, \U) + (1-\pg(\U))\PriorOOD  \notag \\
    \MoveEqLeft[3] p'(y|\S) =  p^{S}(y, \S) + (1-\pg(\S))\PriorinD  
    \label{eq_da_laplace}
\end{align}
}

From a principled probabilistic perspective, \eqref{eq_da_laplace} applies the law of total probability and weigh two terms: (1) The classifier predictions $p(y|\U)$; (2) A uniform smoothing prior $\pi^\U$. They are weighed by the belief that the input image is from a class that is familiar to the expert $p(\U)$ or unfamiliar $(1-p(\U))$ . 
The rightmost term means that \textit{given that we know} an image is unfamiliar for an expert, we assign a uniform low probability, which is weighed by the belief that the input image is indeed unfamiliar.

The resulting model has two interesting properties. First, it reduces hyper-parameter tuning, because prior weights are determined automatically. Second, smoothing adds a constant value to each score, hence it maintains the class that achieves the maximum of each individual expert, but at the same time affects their combined prediction in \eqref{eq_total_prob_models}.

{
\begin{algorithm}[t]
   \caption{\COSMO Inference}
   \label{alg}
\begin{algorithmic}[1]
   \STATE {\bfseries Input:} Image
   \STATE Estimate $p^S(y, \S)$ and $p^{ZS}(y, \U)$ of two experts
   \STATE Estimate $p^{Gate}(\S) = f\big(p^S(y, \S), p^{ZS}(y, \U)\big)$ ; Fig. \ref{fig2}
   \STATE Estimate $p'(y|\S)$ and $p'(y|\U)$ by smoothing; \eqref{eq_da_laplace}
   \STATE Estimate $p(y)$ by soft-combining; \eqref{eq_total_prob_models} 
\end{algorithmic}
\end{algorithm}
}
\section{Details of our approach} 
\label{sec_details}
Our approach has three learning modules: A model for seen classes, for unseen classes, and for telling them apart. 
The three components are trained separately. Supp. section \ref{sec_joint_training} explains why they cannot be trained jointly in this setup. 

\vspace{5pt}
\noindent\textbf{A model for unseen classes.}
For unseen classes, we use either LAGO  \cite{LAGO} or fCLSWGAN \cite{xian_2018} with the code provided by the authors. Each of these models achieves state-of-the-art results for part of ZSL benchmarks. LAGO predicts $p^{ZS}(\y|\x)$ by learning an AND-OR  group structure over attributes.  fCLSWGAN \cite{xian_2018} uses GAN to augment the training data with synthetic feature vectors of unseen classes, and then trains a classifier to recognize the classes. We retrained the models on the GZSL split (Figure \ref{fig_splits}).%

\vspace{5pt}
\noindent\textbf{A model for seen classes.}
For seen classes, we trained a logistic regression classifier to predict $p^{S}(\y|\x)$. We used  a LBFGS solver \cite{LBFGS} with default aggressiveness hyper-parameter (C=1) of sci-kit learn \cite{sklearn}, as it exhibits good performance over the Seen-Val set (Figure \ref{fig_splits}). 

\vspace{5pt}
\noindent\textbf{A confidence-based gating model.}
To discriminate between Seen and Unseen classes, we use a logistic regression classifier to predict $p(\S|\x)$, trained on the \textit{Gating-Train set} (Figure \ref{fig_splits}).
For input features, we use softmax scores of both the \textit{unseen} expert ($p^{ZS}$) and seen expert ($p^S$). We also apply temperature scaling \cite{ODIN} to inputs from $p^S$, \figref{fig2}. 

We used the sci-kit learn LBFGS solver with default aggressiveness hyper parameter (C=1) because the number of weights (\tildeapprox 10-50) is much smaller than the number of training samples (\tildeapprox thousands). 
We tune the decision threshold and softness of the gating model by adding constant bias $\beta$ and applying a sigmoid with $\gamma$ gain on top of its scores:
\begin{equation}
p(\S|\x) = \sigma\Big\{ \gamma [score - \beta] \Big\} \quad.
\label{eq_sigmoid_gating}
\end{equation}
$\gamma$  and $\beta$ were tuned using cross validation.

\ignore{In the supplementary we describe an ablation study for our design decisions for selecting the input features. We observe that (1) in agreement with \cite{ODIN} we observe that increasing the softmax temperature is informative for out-of-distribution detection. (2) Using the unseen class predictions provides an additional discriminative signal.}

\begin{figure*}[htbp]
    \centering
  \includegraphics[height=6.2 cm, trim={0.27cm 0.48cm 0.6cm 0.52cm},clip]{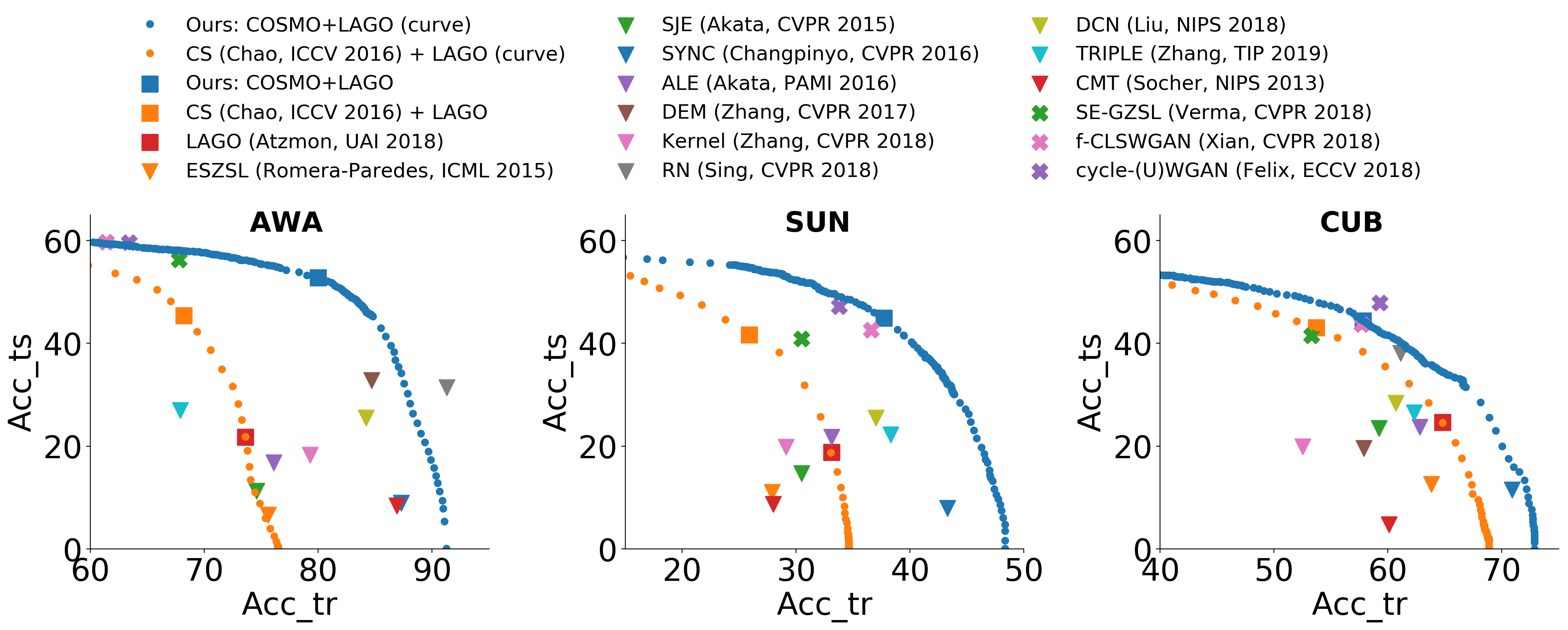} %
  \caption{ The Seen-Unseen curve for \COSMOLAGO\!, compared with: (1) The curve of \CSLAGO \cite{chao} baseline, (2) 15 baseline GZSL models. Dot markers denote samples of each curve. \textbf{Squares}: \COSMO cross-validated model and its LAGO-based baselines. \textbf{Triangles}: non-generative approaches, \textbf{'X'}: approaches based on generative-models.
  Generative models tend to tend to be biased toward the Unseen classes, while non-generative models tend to be biased toward the Seen classes. Importantly, the \COSMO curve achieves a better or equivalent performance compared to all methods, and allows to easily choose any operation point along the curve.}
    \label{fig_curves}
\end{figure*}

\begin{table*}[htbp]
    \begin{center}
      \begin{small}\begin{sc}
      \scalebox{0.98}{ % mahogany
    \renewcommand{\arraystretch}{0.8} %
    \setlength{\tabcolsep}{4pt} %
\begin{tabular}{l|lll|lll|lll|lll}
\toprule
dataset & \multicolumn{3}{l|}{AWA} & \multicolumn{3}{l|}{SUN} & \multicolumn{3}{l|}{CUB} & \multicolumn{3}{l}{FLOWER} \\
{} & $Acc_{ts}$ & $Acc_{tr}$ &        $Acc_H$ & $Acc_{ts}$ & $Acc_{tr}$ &        $Acc_H$ & $Acc_{ts}$ & $Acc_{tr}$ &        $Acc_H$ & $Acc_{ts}$ & $Acc_{tr}$ &        $Acc_H$ \\
\midrule
{\textbf{NON-GENERATIVE   MODELS}}&&&&&&&&&\\%[2mm]  %

ESZSL \cite{ESZSL}  &        6.6 &       75.6 &           12.1 &         11 &       27.9 &           15.8 &       12.6 &       63.8 &             21 &       11.4 &       56.8 &             19 \\
SJE \cite{SJE}            &       11.3 &       74.6 &           19.6 &       14.7 &       30.5 &           19.8 &       23.5 &       59.2 &           33.6 &       13.9 &       47.6 &           21.5 \\
DEVISE  \cite{DEVISE}         &       13.4 &       68.7 &           22.4 &       16.9 &       27.4 &           20.9 &       23.8 &         53 &           32.8 &        9.9 &       44.2 &           16.2 \\
SYNC  \cite{SYNC}      &        8.9 &       87.3 &           16.2 &        7.9 &       43.3 &           13.4 &       11.5 &       70.9 &           19.8 &          - &          - &              - \\
ALE \cite{ALE}        &       16.8 &       76.1 &           27.5 &       21.8 &       33.1 &           26.3 &       23.7 &       62.8 &           34.4 &       34.4 &       13.3 &           21.9 \\
DEM \cite{DEM}            &       32.8 &       84.7 &           47.3 &          - &          - &              - &       19.6 &       57.9 &           29.2 &          - &          - &              - \\
Kernel \cite{zhang_kernel}        &       18.3 &       79.3 &           29.8 &       19.8 &       29.1 &           23.6 &       19.9 &       52.5 &           28.9 &          - &          - &              - \\
ICINESS  \cite{ICINESS}        &          - &          - &              - &          - &          - &           30.3 &          - &          - &           41.8 &          - &          - &              - \\
TRIPLE  \cite{TRIPLE}          &         27 &       67.9 &           38.6 &       22.2 &       38.3 &           28.1 &       26.5 &       62.3 &           37.2 &          - &          - &              - \\
RN \cite{RelationNet}              &       31.4 &       91.3 &           46.7 &          - &          - &              - &       38.1 &       61.1 &             47 &          - &          - &              - \\

\midrule
{\textbf{GENERATIVE MODELS}}&&&&&&&&&\\%[2mm] %
SE-GZSL \cite{CVAE2}      &       56.3 &       67.8 &  61.5 &       40.9 &       30.5 &           34.9 &       41.5 &       53.3 &           46.7 &          - &          - &              - \\
fCLSWGAN \cite{xian_2018}     &       59.7 &       61.4 &           59.6 &       42.6 &       36.6 &  39.4 &       43.7 &       57.7 &           49.7 &         59 &       73.8 &           65.6 \\
fCLSWGAN* (reproduced)                        &       53.6 &         67 &  59.6 &       40.1 &         36 &  37.9 &       45.1 &       55.5 &  49.8 &       58.1 &       73.2 &           64.8 \\
cycle-(U)WGAN  \cite{CCGAN} &       59.6 &       63.4 &           59.8 &       47.2 &       33.8 & 39.4 &       47.9 &       59.3 &  \textbf{53.0} &       61.6 &       69.2 &           65.2 \\

\midrule
{\textbf{\DAZL AND  BASELINES}}&&&&&&&&&\\%[2mm] %

CMT \cite{socher2013zero}          &        8.4 &       86.9 &           15.3 &        8.7 &         28 &           13.3 &        4.7 &       60.1 &            8.7 &          - &          - &              - \\
DCN \cite{DCN}               &       25.5 &       84.2 &           39.1 &       25.5 &         37 &           30.2 &       28.4 &       60.7 &           38.7 &          - &          - &              - \\
LAGO  \cite{LAGO}         &       21.8 &       73.6 &           33.7 &       18.8 &       33.1 &           23.9 &       24.6 &       64.8 &           35.6 &  - & - & - \\
CS \cite{chao} + LAGO      &       45.4 &       68.2 &           54.5 &       41.7 &       25.9 &           31.9 &       43.1 &       53.7 &           47.9 &  - & - & - \\
Ours: COSMO+fCLSWGAN*          &       64.8 &       51.7 &           57.5 &       35.3 &       40.2 &           37.6 &       41.0 &       60.5 &           48.9 &       59.6 &       81.4 &  \textbf{68.8} \\
Ours: \COSMOLAGO               &       52.8 &         80 &  \textbf{63.6} &       44.9 &       37.7 &  \textbf{41.0} &       44.4 &       57.8 &  \textbf{50.2} &  - & - & -  \\

\bottomrule
\end{tabular}
      }
      \end{sc}
      \end{small}
    \end{center}
    \vspace{-10pt}
    \caption{Comparing \dazl with state-of-the-art GZSL non-generative models and with generative models that synthesize feature vectors.
    $Acc_{tr}$ is the accuracy of seen classes, $Acc_{ts}$ is the accuracy of unseen classes and \AccH is their harmonic mean. %
    \COSMOLAGO uses LAGO \cite{LAGO} as a baseline GZSL model, and respectively \COSMOGAN uses fCLSWGAN \cite{xian_2018}. 
    \COSMOLAGO improves \AccH over state-of-the-art models by 34\%, 35\%, 7\% respectively for AWA, SUN and CUB.
    Comparing with generative models, \COSMOLAGO closes the non-generative:generative performance gap, and is comparable to or better than these models, while is very easy to train. }
    \label{table_test_vs_models}
    
\end{table*}

\section{Experiments}
\label{sec_experiments}
We tested \COSMO on four GZSL benchmarks and compared to 17 state-of-the art approaches.

The source code to reproduce our experiments is under \url{http://chechiklab.biu.ac.il/~yuvval/COSMO/}. 

\subsection{Evaluation protocol}
To evaluate \COSMO\!\! we follow the protocol of Xian \cite{xianCVPR,xian_awa2}, which became the common experimental framework for comparing GZSL methods. Our evaluation uses its features (ResNet \cite{ResNet}), cross-validation splits, and evaluation metrics for comparing to  state-of-the-art  baselines.

\vspace{5pt}
\noindent\textbf{Evaluation Metrics:}
By definition, GZSL aims at two different sub-tasks: classify seen classes and classify unseen classes. The standard GZSL evaluation metrics therefore combine accuracy from these two sub-tasks. 
Following \cite{xian_awa2}, we report the harmonic mean of $Acc_{tr}$ - the accuracy over seen classes, and $Acc_{ts}$ - the accuracy over unseen classes, see Eq. 21 in \cite{xian_awa2}, $Acc_H =  2{(Acc_{ts}Acc_{tr)}}/{(Acc_{ts}+Acc_{tr})}$.

As a second metric, we compute the full seen-unseen accuracy curve using a parameter to sweep over the decision threshold.
Like the precision-recall curve or ROC curve, the seen-unseen curve provides a tunable trade-off between the performance over the seen and unseen domains. 

Finally, we report the \textit{Area Under Seen-Unseen Curve} (AUSUC) \cite{chao}.

\subsection{Datasets}
\label{sec_datasets}
We tested \COSMO on four \gzsl benchmark datasets: CUB, AWA, SUN and FLOWER.

\noindent\textbf{CUB} \cite{CUB}\textbf{:}  is a task of fine-grained classification of bird-species. 
CUB has 11,788 images of 200 bird species. Each species described by 312  attributes
(like \textit{wing-color-olive},  \textit{beak-shape-curved}). 
It has 100 seen training classes, 50 unseen validation and 50 unseen test classes. 

\noindent\textbf{AWA:} Animals with Attributes (AWA) \cite{DAP}  consists of 30,475 images of 50 animal classes. Classes and attributes are aligned with the class-attribute matrix of  \cite{Osherson91,Kemp2006}, using a vocabulary of 85  attributes (like \textit{white}, \textit{brown}, \textit{stripes}, \textit{eat-fish}). %
It has 27 seen training classes, 13 unseen validation and 10 unseen test classes. 

\noindent\textbf{SUN} \cite{SUN}\textbf{:} is a dataset of complex visual scenes, having 14,340 images from 717 scene types and 102 semantic attributes. 
It has 580 seen training classes, 65 unseen validation and 72 unseen test classes. 

\noindent\textbf{FLOWER} \cite{FLO}\textbf{:}  is a dataset of fine-grained classification of flowers, with 8189 images of 102 classes. Class descriptions are based on sentence embedding from \cite{Reed}. We did not test \COSMOLAGO with this dataset because LAGO cannot use sentence embedding.

\begin{figure}[ht!]
    \centering
      \includegraphics[height=3.cm, trim={5.8cm 11.cm 7.8cm 3.5cm},clip]{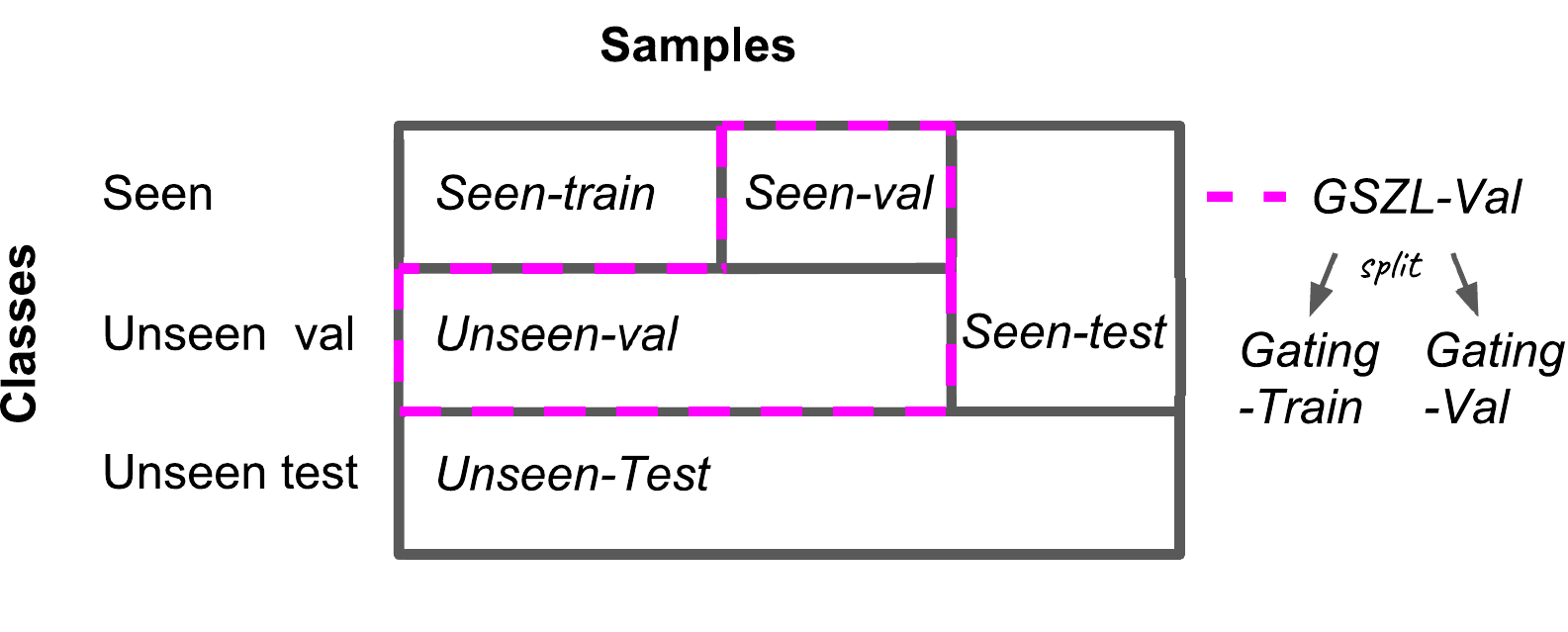} %
      \caption{GZSL cross-validation splits. The data is organized across classes and samples. We define \textbf{Seen-Val} as a subset of the seen-training samples  provided by \cite{xianCVPR, xian_awa2}. We define \textbf{GZSL-Val} = Seen-Val $\cup$ Unseen-Val (in pink). We use GZSL-Val to select the model's  hyper-parameters and learn (\tildeapprox 10-50) weights of the gating network. We split GZSL-Val to \textbf{Gating-Train} and \textbf{Gating-Val} subsets, and use Gating-Train as the held-out subset to train the gating model and  Gating-Val to evaluate its metrics.}
    \label{fig_splits}
\end{figure}

\subsection{Cross-validation}
For selecting hyper-parameters for \COSMO\!\!, we make two additional splits: \textit{GZSL-val} and \textit{Gating Train / Val}. See Figure \ref{fig_splits} for details.

We used cross-validation to optimize the \AccH metric over $\beta$ and $\gamma$ of \eqref{eq_sigmoid_gating} on the \textit{GZSL-Val} set. We tuned these hyper params by first taking a coarse grid search, and then making a finer search around the best performing values for the threshold.
Independently, we used cross-validation on \textit{Gating-Train/Val} to optimize the out-of-distribution AUC over $T$ (Temperature) and $K$ (for top-K pooling).

We stress that training the gating network using \textit{Gating-Train/Val} is not considered as training with external data, because in accordance with \cite{xian_awa2}, once hyper parameters are selected, models are retrained on the union of the training and the validation sets (excluding the gating model). %

\subsection{Compared methods}
We compare \COSMO with 17 leading GZSL methods. These include widely-used baselines like 
\textbf{ESZSL} \cite{ESZSL},
\textbf{ALE} \cite{ALE},
\textbf{SYNC} \cite{SYNC},
\textbf{SJE} \cite{SJE},
\textbf{DEVISE} \cite{DEVISE},
recently published approaches 
\textbf{RN} \cite{RelationNet},
\textbf{DEM} \cite{DEM},
\textbf{ICINESS} \cite{ICINESS},
\textbf{TRIPLE} \cite{TRIPLE},
\textbf{Kernel} \cite{zhang_kernel} 
and methods that provide interesting insight into the method, including 
\textbf{CMT} \cite{socher2013zero}, \textbf{DCN} \cite{DCN},
\textbf{LAGO} \cite{LAGO} and 
\textbf{CS} \cite{chao}, which we reproduced using LAGO as a ZSL module. 

Recent work showed that generating synthetic samples of unseen classes using GANs or VAEs \cite{xian_2018, CCGAN, CVAE2, ZhuGAN} can substantially improve \gzsl. The recent literature considers this generative effort to be orthogonal to modelling, since the two efforts can be combined 
\cite{DCN, LAGO, DIPL_NIPS, ICINESS, TRIPLE}.
Here we compare \COSMO directly both with the approaches listed above, and with generative approaches \textbf{fCLSWGAN} \cite{xian_2018}, \textbf{cycle-(U)WGAN} \cite{CCGAN}, \textbf{SE-GZSL} \cite{CVAE2}. %

\begin{table}[htbp]
    \begin{center}
      \begin{small}\begin{sc}
    \renewcommand{\arraystretch}{0.8} %
    \setlength{\tabcolsep}{4pt} %

\begin{tabular}{lllll}
\toprule
{} &            AWA &            SUN &            CUB &            FLOWER  \\
\midrule

ESZSL         &           39.8 &           12.8 &           30.2 &           25.7  \\
LAGO \cite{LAGO} &          43.4 &           16.3 &           34.3 & -  \\
fCLSWGAN               &           46.1 &           22 &           34.5  &           53.1 \\
cycle-(U)WGAN          &             45 &  22.5 &  \textbf{40.4} &           56.9\\

\midrule
\COSMO \& fCLSWGAN                    &  \textbf{55.9} &             21 &           35.6 &  \textbf{58.1} \\
\COSMO \& LAGO                          & 53.2 &  \textbf{23.9} &           35.7 &  - \\

\bottomrule
\end{tabular}
      \end{sc}\end{small}
    \end{center}
        \vspace{-10pt}
    \caption{Area Under Seen-Unseen Curve (AUSUC)  on the test set: On all datasets, \COSMO improves AUSUC for LAGO and fCLSWGAN. \COSMO introduces new state-of-the-art results on 3 of 4 datasets. 
    }
    \label{table_ausuc}
\end{table}

\section{Results}
\label{sec_results}
We first describe the performance of \COSMO on the test set of the four benchmarks and compare them with baseline methods. We then study in greater depth the properties of \COSMO\!\!, through a series of ablation experiments. 

Table \ref{table_test_vs_models} describes the test accuracy of \COSMOLAGO\!, \COSMOGAN and compared methods over the four benchmark datasets.
Compared with non-generative models, \COSMOLAGO improves the harmonic accuracy \AccH by a large margin for all four datasets: 63.6\% vs 47.3\% in AWA, 41\% vs 30.3\% in SUN and 50.2\% vs 47\% in CUB. 

In addition, \COSMOLAGO closes the performance gap with generative approaches. It wins in AWA (63.6\% versus 61.5\%) and SUN (41\% versus 39.4\%), and loses in CUB (50.2\% versus 53\%). 
Interestingly, \COSMOLAGO reaches state-of-the-art performance although LAGO alone performed poorly on the \textit{generalized} ZSL task. %

\COSMOGAN provides a new state-of-the-art result on FLOWER (68.8\% vs 65.6\%). On AWA, SUN and CUB it achieves lower performance than fCLSWGAN and \COSMOLAGO\!. This happens because the chosen operating point for (\Acctr, \Accts\!), selected by cross validation, was not optimal for the harmonic accuracy \AccH\!. More details are provided in Supplementary \ref{sec_curves_GAN}.

\begin{table*}[htbp]
    \begin{center}
      \begin{small}\begin{sc}
    \renewcommand{\arraystretch}{0.8} %
\begin{tabular}{l|lll|lll|lll}
\toprule
   \ADmodel & \multicolumn{3}{l|}{AWA} & \multicolumn{3}{l|}{SUN} & \multicolumn{3}{l}{CUB} \\
           &          \AccH &            AUC &      FPR &          \AccH &            AUC &      FPR &          \AccH &            AUC &      FPR \\
\midrule

\pmaxO &           52.9 &           86.7 &           67.9 &           38.5 &           60.9 &           92.8 &           43.3 &           74.1 &           82.4 \\
\pmaxT &           53.1 &           88.6 &           56.8 &           38.4 &             61 &           92.3 &           43.6 &           73.4 &           79.6 \\
\topkNoZS &           52.8 &           88.8 &           56.4 &           38.4 &             61 &           92.2 &           43.8 &           74.2 &           80.1 \\
\topkO &           53.9 &           88.9 &           59.1 &           39.8 &           75.5 &  \textbf{77.5} &  \textbf{45.1} &           81.7 &           73.1 \\
\topkT &  \textbf{56.8} &  \textbf{92.5} &  \textbf{45.5} &  \textbf{40.1} &  \textbf{77.7} &  \textbf{77.5} &           44.8 &  \textbf{82.0} &  \textbf{72.0} \\

\bottomrule
\end{tabular}
      \end{sc}\end{small}
    \end{center}
        \vspace{-10pt}
    \caption{Ablation study for various \gating model variants on validation set. AUC denotes Area-Under-Curve when sweeping over detection threshold. FPR denotes  False-Positive-Rate on the threshold that yields 95\% True Positive Rate for detecting in-distribution samples.
    }
    \label{table_ad_val_direct}
\end{table*}

\subsection{The seen-unseen plane} 
By definition, the GZSL task aims to perform well in two different metrics: accuracy on seen classes and on unseen classes. It is therefore natural to compare approaches by their performance on the seen-unseen plane. 
This is important, because different approaches may select different operating-points  to trade seen and unseen accuracy. 

In Figure \ref{fig_curves} we provide a full Seen-Unseen curve (blue dots) that shows how \COSMOLAGO trade-off the metrics. %
We compare it with a curve that we computed for the \CSLAGO baseline (orange dots) and also show the  results (operation-points) reported for the compared methods. For plotting the curves, we sweep over the decision threshold ($\beta$) of the gating network, trading its true-positive-rate with its false-positive-rate. In the blue-square we show our operating point which was selected with cross-validation by choosing the best \AccH on GZSL-Val set.
 
An interesting observation is that different types of models populate different regions of the Seen-Unseen curve. Generative models (X markers) tend to favor unseen-classes accuracy over accuracy of seen classes, while non-generative models (triangles) tend to favor seen classes. Importantly, \COSMO can be tuned to select any operation point along the curve, and achieve better or equivalent performance at all regions of the seen-unseen plane.

In Supplementary Figure \ref{fig_curves_GAN}, we provide the curves of \COSMOGAN for AWA, SUN, CUB and FLOWER.

Table \ref{table_ausuc} reports the AUSUC of \COSMO compared to four  baseline models. To produce the full curve for the baselines, we used the code provided by the authors and applied  calibrated-stacking \cite{chao} with a series of constants. 
On all four datasets, \COSMO improves AUSUC for LAGO and fCLSWGAN. \COSMO also introduces new state-of-the-art AUSUC on AWA, SUN and FLOWER.

\subsection{Ablation experiments}
To understand the contribution of the different modules of \COSMO\!\!, we carried ablation experiments on \COSMOLAGO that quantify the benefits of the CBG network and adaptive smoothing.

We first compared variants of the gating model, then compared variants of the smoothing method, and finally, compared how these modules work together.

\textbf{Confidence-Based \Gating:} Table \ref{table_ad_val_direct} describes: (1) The OOD metrics \textit{AUC} and \textit{False-Positive-Rate at 95\% True-Positive-Rate} on \textit{Gating-Val}. (2) \AccH metric on \textit{GZSL-Val}. %

We test the effect of temperature scaling and of confidence-based gating by comparing the following gating models: 
(1) \textit{\topkT} is our best confidence-based \gating model, from  \secref{sec_domain_aware_gating} with temperature $T=3$. 
(2) \textit{\topkO} is the same model with $T=1$, revealing the effect of temperature scaling \cite{ODIN}.
(3) \textit{\topkNoZS} is like \textit{\topkT} without the inputs from the ZS expert, revealing the importance of utilizing  information from both experts.
(4) \textit{\pmaxO} is a baseline gating model of \cite{Hendrycks}, instead of the CBG network, it classifies $\S/\U$ by comparing the largest  softmax score to a threshold. 
(5) \textit{\pmaxT} is like \textit{\pmaxO}, but with $T=3$.  
In these experiments, smoothing is disabled to only quantify factors related to the \gating model.

We find that both temperature scaling and confidence-based gating improves the quality metrics. Importantly, confidence-based gating has a strong contribution to performance: The AUC increases from 86.7 to 92.5 for AWA, 60.9 to 77.7 for SUN and 74.1 to 82 for CUB.

\textbf{Adaptive Confidence Smoothing:}
Table \ref{table_ablation_direct} shows the contribution of adaptive smoothing to \AccH on the validation set.
In these experiments, \gating is disabled, to only quantify factors related to the smoothing. (1) \textit{\laplaceAd} corresponds to \eqref{eq_da_laplace}. (2) \textit{\laplaceConst} uses a constant smoothing weight $\lambda$ (Eq. \ref{eq_da_laplace_lambda}) for all images. $\lambda$ was selected by cross validation. \textit{\laplaceAd} shows superior performance to \textit{\laplaceConst} (AWA: 54.3\% vs 53\%, SUN: 41.4\% vs 38.4\%, CUB: 45.7\% vs 43.6\%) and even  to not using smoothing at all ($\lambda=0$).

\begin{table}[htbp]
    \begin{center}
      \begin{small}\begin{sc}
        \renewcommand{\arraystretch}{0.8} %
    \begin{tabular}{llll}
\toprule
               &           AWA &            SUN &            CUB \\
\midrule

   $\lambda=0$ &           52.9 &           38.4 &           43.4 \\
 \laplaceConst &             53 &           38.6 &           43.6 \\
     \laplaceAd &  \textbf{54.3} &  \textbf{41.4} &  \textbf{45.7} \\
\bottomrule %
\end{tabular}

 \end{sc}\end{small}
    \end{center}
        \vspace{-10pt}
    \caption{Ablation study for adaptive smoothing. Showing \AccH on GZSL-Val set.}
    \label{table_ablation_direct}
\end{table}

\vspace{-10pt}

\begin{table}[htbp]
    \begin{center}
      \begin{small}\begin{sc}
    \renewcommand{\arraystretch}{0.8} %
\begin{tabular}{llll}
\toprule
{} &            AWA &            SUN &            CUB \\
\midrule

Independent-Hard           &           58.3 &           35.1 &           44.6 \\
Independent-Soft           &           57.7 &           37.3 &           46.8 \\
CB-Gating                &  \textbf{64.3} &           39.7 &           49.1 \\
Adaptive-Smoothing             &           63.6 &           40.8 &           49.6 \\
\COSMO (Gating \& Smoothing) &           63.6 &  \textbf{41.0} &  \textbf{50.2} \\

\bottomrule
\end{tabular}
\vspace{-10pt}
      \end{sc}\end{small}
    \end{center}
    \caption{Ablation study for combining smoothing and gating, showing \AccH on the test set.}
\label{table_ablation_test_direct}

\end{table}

\noindent\textbf{Combining gating and smoothing:} Table \ref{table_ablation_test_direct} reports test-\AccH when ablating the main modules of \COSMO:

(1) \textit{Independent-Hard} is the simplest approach, applying \eqref{eq_total_prob_models} where the modules don't exchange information, and the gating is a hard decision over ``Max of Softmax" \cite{Hendrycks}. It resembles reproducing CMT \cite{socher2013zero} but using LAGO as the ZSL model and Max-of-softmax gate.
(2) \textit{Independent-Soft} uses a soft gating following  \eqref{eq_sigmoid_gating}.
(3) \textit{CB-Gating} applies the CBG network (\secref{sec_domain_aware_gating}) for the gating module.
(4) \textit{Adaptive-Smoothing} uses \eqref{eq_da_laplace}, and max-of-softmax gating. 
(5) \textit{\COSMO} is our best model, applying both CB-Gating and adaptive smoothing.

We find that both adaptive confidence smoothing and confidence-based gating contribute to test accuracy. Compared with Independent-Hard, \COSMO shows a relative improvement from 58.3\% to 63.6\% on AWA, 35.1\% to 41\% on SUN and 44.6\% to 50.2\% on CUB. Adaptive confidence smoothing and confidence-based gating are weakly synergistic, providing a 1.2\% relative improvement for CUB, 0.5\% for SUN and -1.1\% for AWA.

Importantly, accuracy of \COSMO is comparable with  state-of-the-art generative models. This is important because \COSMO is much easier to train and tune than GAN-based approaches.

\textbf{Note about results:} On the final CVPR version of this paper, we made a typing error in Table \ref{table_ad_val_direct}, mixing results of \COSMOLAGO with  \COSMOGAN\!\!. The results reported here are the correct ones.

\textbf{Acknowledgements:} 
The Titan Xp used for this research was donated by the NVIDIA Corporation. This research was supported by an Israel science foundation grant 737/18.

{\small
\bibliographystyle{ieee_fullname}
\bibliography{egbib}
}

\clearpage
\newpage

\title{Supplementary Material}
\author{}
\onecolumn

\renewcommand\thefigure{S.\arabic{figure}}    
\renewcommand\thetable{S.\arabic{table}}   
\renewcommand\thesection{\Alph{section}}   
\renewcommand{\theequation}{S.\arabic{equation}}
\setcounter{figure}{0}  
\setcounter{table}{0}  
\setcounter{section}{0}  

\maketitle
\section{A walk-through example}
\label{sec_walkthrough_example}

\begin{figure*}[h]
    \centering
  \includegraphics[height=7.cm, trim={2.cm 2cm 1.cm 0.5cm},clip]{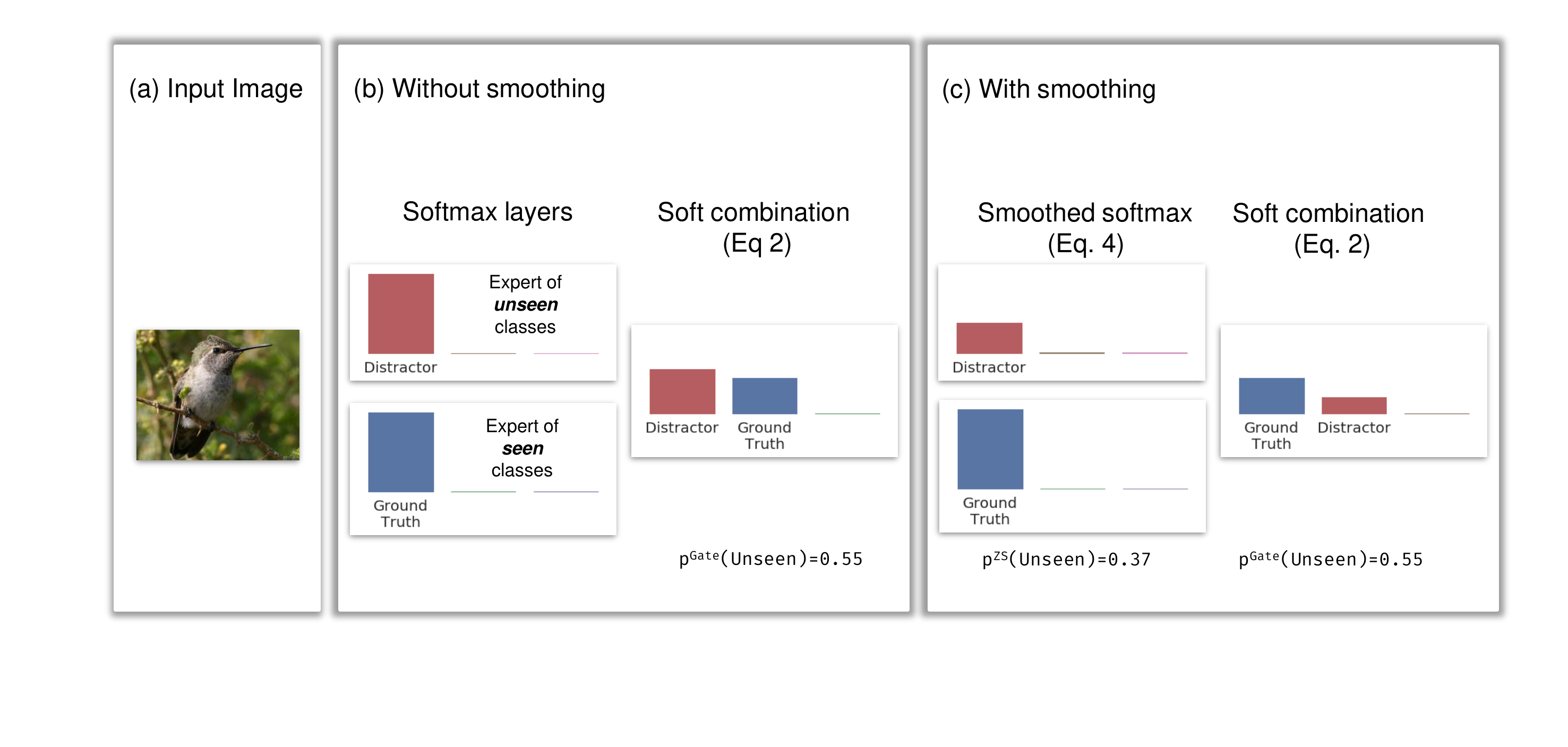} %
  \caption{ A walk-through example  }
    \label{fig_walkthrough}
\end{figure*}

\figref{fig_walkthrough} demonstrates the inference process of \COSMO with an without smoothing. An image (\textit{panel a}) is processed by two experts: (1) An expert of unseen classes produces a distribution of confidence scores $p^{ZS}(y,\U)$ (2) An expert of seen classes produces a distribution of confidence scores $p^S(y,\S)$. Next, the CBG gating network (\secref{sec_domain_aware_gating}) combines these confidence scores into a belief $p^{Gate}(\U)$. 

\textit{Without smoothing} (\textit{panel b}): Here, $p^{ZS}(y,\U)$ and $p^{S}(y,\S)$ are normalized to $p^{ZS}(y|\U)$, $p^{S}(y|\S)$ and then a joint prediction is estimated by soft combining the modules with \eqref{eq_total_prob_models}. In the example, the unseen expert produces overly confident prediction for a wrong (\textit{distractor}) class (red bar). When soft combining the expert decisions, this overwhelms the correct decision of the seen expert (blue bar), producing a false positive detection of distractor class.

\textit{With smoothing} (\textit{panel c}): Here, $p^{ZS}(y,\U)$ and $p^{S}(y,\S)$ are smoothed to $p'(y|\U)$, $p'(y|\S)$ with \eqref{eq_da_laplace} and then a joint prediction is estimated by soft combining the modules with \eqref{eq_total_prob_models}. In the example, the over confident prediction of the unseen expert is smoothed (red bar). When soft combining the expert decisions, it allows the model to reach a correct decision (blue bar).

\section{Negative results for OOD methods}
\label{sec_perturbation_results}
We tested two state-of-the-art methods for out-of-distribution detection:  \textit{ODIN} \cite{ODIN} and \textit{Ensemble} (Ensemble, \cite{Vyas}). We observed that taking a perturbation hurts OOD metrics with both these methods. In addition, in \textit{Ensemble}, although quality metrics improved for the left-out training subsets during training time, the ensemble models learned to overfit the left-out subsets and failed to generalize to Unseen-Val set, better than using the baseline \pmaxO\!.

We believe this result may be due to two factors: \textbf{ (1) Fine-grained datasets are harder:} CUB, SUN and AWA are fine grained datasets. For an un-trained eye, all their unseen samples may appear as in-distribution. For example, only a few fine-grained details discriminate ``Black Throated Blue Warbler'' ($\in \S$) of ``Cerulean Warbler'' ($\in \U$). Therefore we believe that a perturbation would have a similar effect on images from $\S$ or $\U$. \textbf{(2) Shallow vs Deep:} In the standard GZSL protocol we use, each sample is represented as a feature vector extracted from a deep CNN pre-trained on ImageNet. We found that the best classifier for this data is a shallow logistic regression classifier. This is different than ODIN and Ensemble that make the perturbation along a deep network.

\section{Seen-Unseen curves for \COSMOGAN \cite{xian_2018}}
\label{sec_curves_GAN}

\begin{figure*}[h]
    \centering
  \includegraphics[height=6.2cm, trim={0.cm 0cm 0cm 0cm},clip]{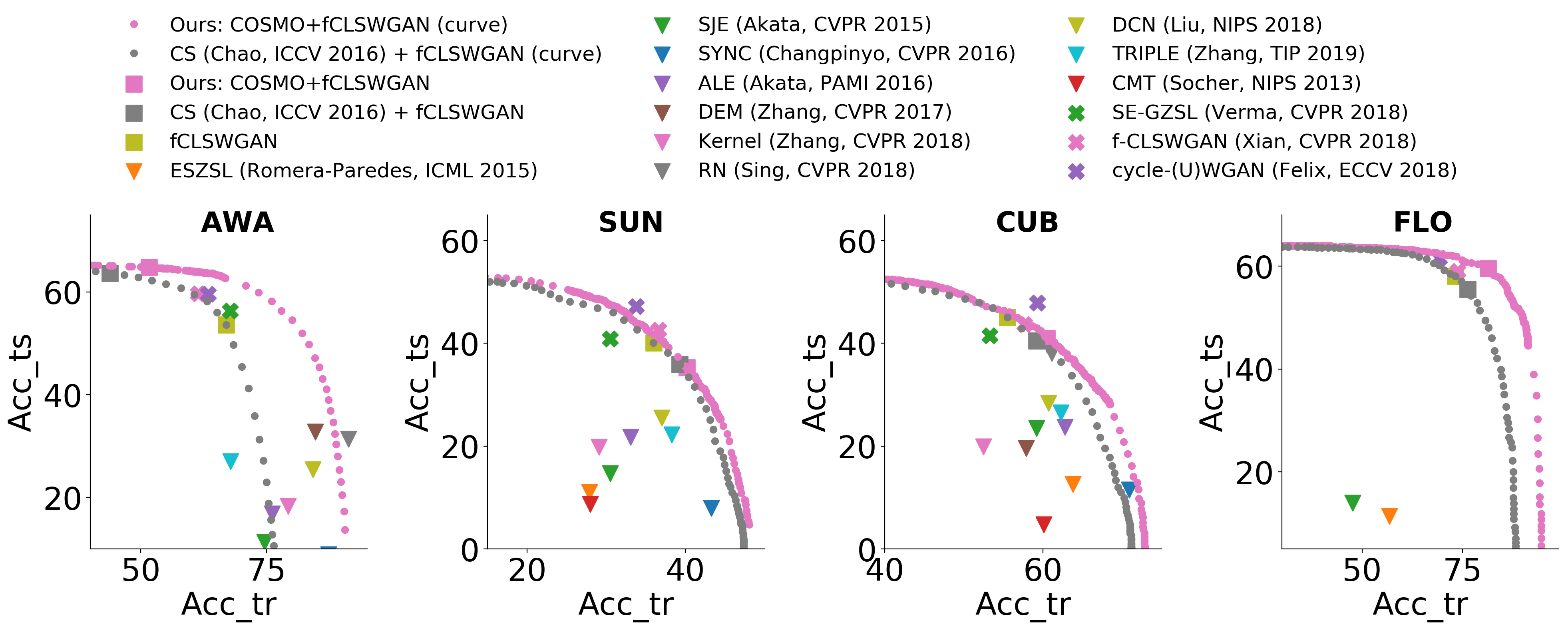} %
  \caption{ The Seen-Unseen curve for \COSMOGAN, compared to: (1) The curve of \textit{CS \cite{chao} +fCLSWGAN}  baseline, (2) 15 baseline GZSL models. Dot markers denote samples of each curve. \textbf{Squares}: \dazl cross-validated model and its fCLSWGAN*-based baselines. \textbf{Triangles}: non-generative approaches, \textbf{'X'}: approaches based on generative-models.}
    \label{fig_curves_GAN}
\end{figure*}

Figure \ref{fig_curves_GAN} provides a full Seen-Unseen curve (pink dots) that shows how \COSMOGAN trades-off the metrics.
We compare it with a curve that we computed for the CS+fCLSWGAN baseline (gray dots) and also show the  results (operation-points) reported for the compared methods (pink-square), selected with cross-validation by choosing the best \AccH on GZSL-Val set.

The pink curve shows that on all datasets, \COSMO produces equivalent or better performance compared to fCLSWGAN baseline (pink-\textbf{X}). However in most cases the operation-point selected with cross validation (pink-square) is inferior to fCLSWGAN baseline (olive-square and Table \ref{table_test_vs_models}).

\section{Joint training of all modules}
\label{sec_joint_training}
We now explain why the GZSL setup prevents from training the gater jointly with the $\S$ and $\U$ experts.  
Basiclaly, in GZSL, one cannot mix seen and unseen samples during the same learning phase. More specifically, to adhere to the standard GZSL protocol by \cite{xianCVPR} in which some test samples come from unseen validation classes, one has two options. (1) Do not use these  classes when training the seen expert S. This decimates S's accuracy on them. (2) Do use them for training S. In that case, all labeled samples are “seen” and the \textit{gater}  cannot learn to discriminate seen from unseen.

We ran two experiments on CUB to evaluate these two options, training the components jointly with a unified loss. 

In the first case, accuracy on seen classes degrades from 72.8\% to 53\%, and on the GZSL task \AccH degrades from 50.2\% (\COSMO) to 26.3\%. 
In the second case, there were no samples of unseen classes when training the model. This greatly hurts the accuracy, leading to: \Accts = 0.1\%, \Acctr = 72.8\%, \AccH=1.9\% far worse than original COSMO.

\end{document}